\documentclass[final]{l4dc2020} 

\makeatletter
\def\set@curr@file#1{\def\@curr@file{#1}} 
\makeatother

\title[Encoding Physical Constraints in Differentiable Newton-Euler Algorithm]{Encoding Physical Constraints in\\Differentiable Newton-Euler Algorithm}
\usepackage{times}

\author{%
 \Name{Giovanni Sutanto\nametag{\thanks{This work was done when Giovanni Sutanto was an intern at Facebook Artificial Intelligence Research.}}} \Email{gsutanto@usc.edu}\\
 \addr Department of Computer Science, University of Southern California
 \AND
 \Name{Austin S. Wang} \Email{wangaustin@fb.com}\\
 \Name{Yixin Lin} \Email{yixinlin@fb.com}\\
 \Name{Mustafa Mukadam} \Email{mukadam@fb.com}\\
 \addr Facebook Artificial Intelligence Research
 \AND
 \Name{Gaurav S. Sukhatme} \Email{gaurav@usc.edu}\\
 \addr Department of Computer Science, University of Southern California
 \AND
 \Name{Akshara Rai} \Email{akshararai@fb.com}\\
 \Name{Franziska Meier} \Email{fmeier@fb.com}\\
 \addr Facebook Artificial Intelligence Research
}

\usepackage{amsmath} 
\usepackage{amssymb}  
\usepackage{algorithm}
\usepackage[noend]{algpseudocode}

\usepackage[shortlabels]{enumitem}
\usepackage{multirow}
\usepackage{booktabs}

\usepackage{url,graphicx,array,mathrsfs,comment}
\usepackage{makeidx} 

\usepackage{verbatim,epstopdf,bm}
\usepackage{float}
\floatstyle{plaintop}
\restylefloat{table}
\usepackage{pgfplots}
\pgfplotsset{compat=newest}
\usepackage{pgfplotstable}
\pgfplotsset{every axis/.append style={font=\footnotesize}}
\usepackage{filemod}
\usepackage{wrapfig}

\usepackage{csquotes}

\usepackage[english]{babel}

\usepackage[disable]{todonotes}

\newcommand{\norm}[1]{\left\lVert#1\right\rVert}
\newcommand{\jointpositionscalar}{q}
\newcommand{\jointposition}{\mathbf{\jointpositionscalar}}
\newcommand{\jointvelocity}{\dot{\jointposition}}
\newcommand{\jointacceleration}{\ddot{\jointposition}}
\newcommand{\cartesianposition}{\mathbf{x}}
\newcommand{\cartesianvelocity}{\dot{\mathbf{x}}}

\newcommand{\eye}{\mathbf{I}}

\newcommand{\jointtorquescalar}{\tau}
\newcommand{\jointtorque}{\boldsymbol{\jointtorquescalar}}

\newcommand{\jointpositiont}{\jointposition_{t}}
\newcommand{\jointvelocityt}{\jointvelocity_{t}}
\newcommand{\jointaccelerationt}{\jointacceleration_{t}}

\newcommand{\jointtorquet}{\jointtorque_{t}}

\newcommand{\inversemodel}{\mathbf{f}_{ID}(\jointposition, \jointvelocity, \jointacceleration)}

\newcommand{\nemodelt}{\mathbf{f}_{NE}(\jointpositiont, \jointvelocityt, \jointaccelerationt; \inertialparams)}

\newcommand{\inversemodelloss}{L_{ID}}
\newcommand{\inertiamatrixscalar}{H}
\newcommand{\inertiamatrix}{\mathbf{\inertiamatrixscalar}}
\newcommand{\inertiamatrixfunction}{\inertiamatrix(\jointposition)}

\newcommand{\coriolismatrixscalar}{C}
\newcommand{\coriolismatrix}{\mathbf{\coriolismatrixscalar}(\jointposition, \jointvelocity)}

\newcommand{\gravityforce}{\mathbf{g}(\jointposition)}

\newcommand{\lowertriangularmatrix}{\mathbf{L}}
\newcommand{\positivedefinitebias}{b}

\newcommand{\rotationmatrix}{\boldsymbol{R}}

\newcommand{\mass}{m}
\newcommand{\centerofmass}{\boldsymbol{c}}
\newcommand{\rotationalinertia}{{\boldsymbol{I}_{C}}}

\newcommand{\zerovector}{\boldsymbol{0}}

\newcommand{\masscenterofmasscombined}{\boldsymbol{h}}

\newcommand{\masscenterofmasscombinedlearnableparam}{\boldsymbol{\theta}_{\masscenterofmasscombined}}

\newcommand{\masslearnableparam}{\theta_{\mass}}

\newcommand{\rotationalinertialearnableparamscalar}{{\theta_{\rotationalinertia}}}

\newcommand{\principalmomentofinertiascalar}{J}
\newcommand{\principalmomentofinertiamatrix}{\boldsymbol{\principalmomentofinertiascalar}}

\newcommand{\principalmomentofinertiaangle}{\alpha}

\newcommand{\densityweightedcovariancerotinertiamatrix}{\boldsymbol{\Sigma}_C}

\newcommand{\inertialparams}{\boldsymbol{\theta}}
\newcommand{\rotmataxisanglelearnableparams}{\boldsymbol{\theta}_\text{RAA}}
\newcommand{\rotmataxisanglelearnableparamscalar}{{\theta_\text{RAA}}}
\newcommand{\triangprmlearnableparams}{\boldsymbol{\theta}_\text{TRI}}
\newcommand{\spdprmlearnableparams}{\boldsymbol{\theta}_\text{SPD}}
\newcommand{\covprmlearnableparams}{\boldsymbol{\theta}_\text{COV}}

\renewcommand{\Re}{\mathbb{R}}

\newcommand{\T}{^{\textrm T}} 

\newcommand{\GSutanto}[1]{\todo[inline,color=blue!40]{GSutanto: #1}}

\newcommand{\FM}[1]{\todo[inline,color=red!40]{FMeier: #1}}

\newcommand{\AW}[1]{\todo[inline,color=pink!40]{Austin: #1}}

\begin{document}

\maketitle

\GSutanto{Can we make the author listing more space-efficient?}

\begin{abstract}%
%
The recursive Newton-Euler Algorithm (RNEA) is a popular technique for computing the dynamics of robots. RNEA can be framed as a differentiable computational graph, enabling the dynamics parameters of the robot to be learned from data via modern auto-differentiation toolboxes. However, the dynamics parameters learned in this manner can be physically implausible. In this work, we incorporate physical constraints in the learning by adding structure to the learned parameters. This results in a framework that can learn physically plausible dynamics via gradient descent, improving the training speed as well as generalization of the learned dynamics models. We evaluate our method on real-time inverse dynamics control tasks on a 7 degree of freedom robot arm, both in simulation and on the real robot. Our experiments study a spectrum of structure added to the parameters of the differentiable RNEA algorithm, and compare their performance and generalization. The code is available at \href{https://github.com/facebookresearch/differentiable-robot-model}{https://github.com/facebookresearch/differentiable-robot-model}.

\end{abstract}

\begin{keywords}%
    learning, structure, rigid body parameters, differentiable, recursive Newton-Euler algorithm, inverse dynamics%
\end{keywords}

\section{Introduction}
\label{sec:learn_inv_dyn_introduction}
%

An accurate dynamics model is key to compliant force control of robots, and there is a rich history of learning of such models for robotics \citep{An_1988_model_based_control, Murray_AMI_1994, atkeson_1986_inertial_param_est}. With an accurate dynamics model, inverse dynamics can be used as a policy to predict the torques required to achieve a desired joint acceleration, given the state of the robot \citep{Murray_AMI_1994}. 

Due to their widespread utility, robot dynamics have been learned in many ways. 
One way is to use a purely data-driven approach with parametric models \citep{Hitzler_LearnAdaptInvDynModels_Humanoids_2019}, non-parametric models \citep{NguyenTuong_LearningInvDynComparison_ESANN08}, and learning error-models \citep{Kappler_NewDataSourceInvDynLearning_IROS17}, in a supervised or self-supervised fashion. However, these purely data-driven approaches typically suffer from a lack of generalization to previously unexplored parts of the state space.
Alternatively, \cite{atkeson_1986_inertial_param_est} recast the dynamics equations such that inertial parameters are a linear function of state-dependent quantities, given the joint torques. 
While inferior to unstructured approaches in terms of flexibility to fit data, this approach typically provides superior generalization capabilities. Recently, \cite{Lutter_delan_iclr19_arxiv, Gupta_structured_learning_mech_sys_arxiv} learn the parameters of Lagrangian dynamics, incorporating benefits of flexible function approximation into structured models. However, these approaches ignore some of the physical relationships and constraints on the learned parameters, which can lead to physically implausible dynamics.

In this work, we combine the modern approach of parameter learning with the structured approach to inverse dynamics learning, similar to \cite{Ledezma_FOPnet_Humanoids17}. We implement the Recursive Newton-Euler algorithm in PyTorch \citep{Paszke_PyTorch_AutoDiff_2017}, allowing the inertial parameters to be learned using gradient descent and automatic-differentiation for gradient computation. The benefit, over traditional least-squares implementations, is that we can easily explore various re-parameterizations of the inertial parameters and additional constraints that encode physical consistency, such as \cite{Wensing_LMI_2018}, without the need for linearity. 
%

In that context, the contribution of this paper is threefold: 
first, we present several re-parameteri-zations of the inertial parameters, which allow us to learn physically plausible parameters using a differentiable recursive Newton-Euler algorithm. 
Second, we show that these re-parameterizations help in improving training speed as well as generalization capability of the model to unseen situations. 
Third, we evaluate a spectrum of structured dynamics learning approaches on a simulated and real 7 degree-of-freedom robot manipulator.

Our results show that adding such structure to the learning can improve the learning speed as well as generalization abilities of the dynamics model. Our models can generalize with much lesser data, and need much fewer training epochs to converge. With our learned dynamics, we see reduced contributions of feedback terms in control, resulting in more compliant motions. 
%

\section{Background and Related Work}
\label{sec:learn_inv_dyn_related_work}
 The dynamics of a robot manipulator are a function of the joint torque $\jointtorque$, joint acceleration $\jointacceleration$ and joint position and velocity $\jointposition, \jointvelocity$:
\begin{equation}
    \jointtorque = \inversemodel = \inertiamatrixfunction \jointacceleration + \coriolismatrix \jointvelocity + \gravityforce
\end{equation}
with $\inertiamatrixfunction, \coriolismatrix, \gravityforce$ are the system inertia matrix, Coriolis matrix, and gravity force, respectively. $\inversemodel$ is the inverse dynamics model that returns the torques that can achieve a desired joint acceleration, given the current joint positions and velocities.

The Recursive Newton-Euler Algorithm (RNEA) \citep{Luh_RNEA_1980, Featherstone_RBDalgo_2007} is a computationally efficient method of computing the inverse dynamics, which scales linearly with respect to the number of the degrees-of-freedom of the robot.


\subsection{Learning models for model-based control}

Accurate inverse dynamics models are crucial for compliant force controlled robots, and hence widely studied in robotics. Previously, researchers have used unstructured multi-layer perceptron (MLP) to learn the complete inverse dynamics \citep{Jansen_InvDynNN_ICANN94, Hitzler_LearnAdaptInvDynModels_Humanoids_2019} or a residual component of the inverse dynamics \citep{Kappler_NewDataSourceInvDynLearning_IROS17}. \cite{NguyenTuong_LearningInvDynComparison_ESANN08} compare non-parametric methods like locally weighted projection regression (LWPR), support vector regression (SVR) and Gaussian processes regression (GPR) for learning inverse dynamics models.


Recently \cite{Lutter_delan_iclr19_arxiv} and concurrently \cite{Gupta_structured_learning_mech_sys_arxiv, Gupta_StructuredMechModels_L4DC20}, proposed a semi-structured learning method for the Lagrangian dynamics of a manipulator, called Deep Lagrangian Networks (DeLaN).
In DeLaN, some of the physical constraints of Lagrangian dynamics are obeyed. For example, the inertia matrix is parametrized to be symmetric positive definite. Moreover, the relationship between coriolis and centrifugal terms and the inertia matrix and joint velocities \citep{Murray_AMI_1994} is satisfied via automatic differentiation. Similarly, the gravity term is derived from a neural network which takes generalized coordinates as input, representing the potential energy. However, other constraints in the dynamics, such as the triangle inequality in the principal moments \citep{Traversaro_IdPhysicalConsistent_2016} of the inertia matrix are not considered. Moreover, neural networks can be sensitive to the chosen architecture and need variations in input data to generalize to new situations.
%
%
Similar to DeLaN, Hamiltonian Neural Networks (HNN) \citep{Greydanus_HamiltonianNN} predict the Hamiltonian (instead of Lagrangian) of a dynamical system. 

Many previous works in parameter identification boil down to setting up a least square problem with some (hard) constraints \citep{Wensing_LMI_2018, Mistry_ParamId_Humanoids09, kozlowski2012modelling}, followed by solving a convex optimization problem. On the other hand, our method incorporates hard constraints as structure in learning representations of the parameters, and then performs back-propagation on the computational graph for optimization. As a result, it is not limited to learning linear parameters, and can generalize to a larger range of problems.
Moreover, our approach can be applied to an online learning setup: for example when the robot carries an additional mass (an object) on one of its links, our approach can adapt the dynamics parameters online, as the robot continues to operate and the data is collected in batches. Traditionally, online learning approaches --including adaptive control \citep{Slotine_AdaptiveControl_IJRR1987}-- do not guarantee physical plausibility of the learned dynamics parameters. On the other hand, more modern system identification methods which incorporate hard constraints on physical parameters \citep{Wensing_LMI_2018, Mistry_ParamId_Humanoids09, kozlowski2012modelling} require collecting data in the new setting before optimizing the new dynamics parameters. 


Our work is closely related to the work by \cite{Ledezma_FOPnet_Humanoids17}, in the sense that our work is also derived from the Newton-Euler formulation of inverse dynamics. However, in our work, we emphasize more on how incorporating structure in learning dynamics parameters helps with improving the training speed and generalization capability of the model. Moreover, we also compare our method with the state-of-the-art semi-structured DeLaN and an unstructured MLP.
\section{Encoding Physical Consistency in Newton-Euler Equations}
\label{sec:learn_inv_dyn_structured_learning_rep_diff_rnea}
The Newton-Euler equations can be implemented as a differentiable computational graph, e.g. with PyTorch \citep{Paszke_PyTorch_AutoDiff_2017}, which we call differentiable NEA (DiffNEA). The parameters of DiffNEA, e.g. the inertial parameters $\inertialparams$, can now be optimized via gradient descent utilizing automatic differentiation to compute the gradients. Although both kinematics and dynamics parameters are involved in Newton-Euler algorithm, in this paper we study the optimization of only the dynamics parameters, assuming the kinematics specification of the robot is correct and fixed.  
\AW{Explain use of boldface and capital letters in notations of scalars, vectors, matrices, and functions?}

Specifically, we aim to optimize parameters $\inertialparams$ of the Newton-Euler equations, such that the inverse dynamics loss is minimized
\begin{equation}
    \inversemodelloss = \sum_{t=1}^{T} \norm{\jointtorquet - \nemodelt}_2^2.
\end{equation}
Typically, $\inertialparams$ is a collection of inertial parameters $\inertialparams_i = [m, \masscenterofmasscombinedlearnableparam, I_{xx}, I_{xy}, I_{xz}, I_{yy}, I_{yz}, I_{zz}]^T \in \Re^{10}$ per link, where $m$ is the link mass, $\masscenterofmasscombinedlearnableparam = [h_x, h_y, h_z] = m \centerofmass$ with $\centerofmass$ being the CoM, and the last 6 parameters representing the rotational inertia matrix $\rotationalinertia$ \citep{atkeson_1986_inertial_param_est}.
%
\AW{The logic behind parameterizing with $h=mc$ might not be clear to people not that familiar with RNEA. Maybe briefly explain or explicitly point the reader to the citation for the explanation? Also you should mention that $\mathbf{\theta_h}$ = $\mathbf{h}$}
When optimizing $\inertialparams$ physical consistency of the estimated parameters is not guaranteed. Enforcing physical constraints on the parameters can be done through explicit constraints \citep{Traversaro_IdPhysicalConsistent_2016, Wensing_LMI_2018}, which requires constrained optimization algorithms to find a solution. 
In the following, we discuss and propose several possible parameter representations $\inertialparams$, which encode increasingly more physical consistency implicitly and allows us to perform unconstrained gradient descent.

\subsection{Unstructured Mass and Rotational Inertia Matrix (DiffNEA No Str)}
We start out with the simplest representation, with an \textit{unconstrained} mass value $\masslearnableparam$ and $9$ \textit{unconstrained} parameters for the rotational inertia matrix: 
\AW{You only mentioned $\theta_m$ and $\theta_{I_{C1-9}}$What about $\mathbf{\theta_h}$?}
\begin{equation}
    \boldsymbol{\theta}_\text{No Str} = 
    \left[
        \masslearnableparam \;\;
        \masscenterofmasscombinedlearnableparam \;\;
        \rotationalinertialearnableparamscalar_{1} \;\; \rotationalinertialearnableparamscalar_{2} \;\; \rotationalinertialearnableparamscalar_{3} \;\; \rotationalinertialearnableparamscalar_{4} \;\; \rotationalinertialearnableparamscalar_{5} \;\;
        \rotationalinertialearnableparamscalar_{6} \;\;
        \rotationalinertialearnableparamscalar_{7} \;\;
        \rotationalinertialearnableparamscalar_{8} \;\;
        \rotationalinertialearnableparamscalar_{9} 
    \right]
    \label{eq:rotinertialearnableparams1}
\end{equation}
This parametrization does not encode any physical constraints and only serves as baseline.
%
\subsection{Symmetric Rotational Inertia Matrix (DiffNEA Symm)}
In this parametrization, we explicitly construct the rotational matrix as a symmetric matrix, with only 6 learnable parameters.
Furthermore, we represent the link mass as: $\mass = (\theta_{\sqrt{m}})^2 + \positivedefinitebias$, where $\positivedefinitebias > 0$ is a (non-learnable) small positive constant to ensure $\mass > 0$. Thus the learnable parameters of this representation are
 \begin{equation}
        \boldsymbol{\theta}_\text{Symm} = 
        \begin{bmatrix}
            \theta_{\sqrt{m}} &
            \masscenterofmasscombinedlearnableparam &
            \rotationalinertialearnableparamscalar_{1} & \rotationalinertialearnableparamscalar_{2} & \rotationalinertialearnableparamscalar_{3} & \rotationalinertialearnableparamscalar_{4} & \rotationalinertialearnableparamscalar_{5} &
            \rotationalinertialearnableparamscalar_{6}
        \end{bmatrix}
        \label{eq:rotinertialearnableparams2}
    \end{equation}
This parameter representation enforces positive mass estimates and symmetric --but not necessarily positive definite-- rotational inertia matrices.
\subsection{Symmetric Positive Definite Rotational Inertia Matrix (DiffNEA SPD)}
Next, we introduce a change of variables, to enforce positive definiteness of the rotational inertia matrix. We construct the lower triangular matrix:
    \begin{equation}
        \lowertriangularmatrix = 
        \begin{bmatrix}
            \theta_{\text{LI}_1} & 0 & 0\\
            \theta_{\text{LI}_4} & \theta_{\text{LI}_2} & 0 \\
            \theta_{\text{LI}_5} & \theta_{\text{LI}_6} & \theta_{\text{LI}_3}\\
        \end{bmatrix}
        \label{eq:parameterized_lower_triang_matrix}
    \end{equation}
    and construct the rotational inertia matrix $\rotationalinertia$ via Cholesky decomposition plus a small positive bias on the diagonal: 
    $\rotationalinertia = \lowertriangularmatrix \lowertriangularmatrix\T + b \eye_{3 \times 3}$. 
    $\eye_{3 \times 3}$ is a $3 \times 3$ identity matrix and $\positivedefinitebias > 0$ is a (non-learnable) small positive constant to ensure positive definiteness of $\rotationalinertia$.
The learnable parameters are:
 \begin{equation}
        \spdprmlearnableparams = 
        \begin{bmatrix}
            \theta_{\sqrt{m}} &
            \masscenterofmasscombinedlearnableparam &
            \theta_{\text{LI}_1} &
            \theta_{\text{LI}_2} &
            \theta_{\text{LI}_3} &
            \theta_{\text{LI}_4} &
            \theta_{\text{LI}_5} &
            \theta_{\text{LI}_6} 
        \end{bmatrix}
        \label{eq:rotinertialearnableparams3}
    \end{equation}
While this representation enforces positive mass and positive definite inertia matrices, it could still lead to inertia estimates that are not physically plausible, as discussed in \cite{Traversaro_IdPhysicalConsistent_2016}. To achieve full consistency, the estimated inertia matrix also needs to fulfill the triangular inequality of the principal moments of inertia of the 3D inertia matrices \citep{Traversaro_IdPhysicalConsistent_2016}. 

\subsection{Triangular Parameterized Rotational Inertia Matrix (DiffNEA Tri)}
To encode the triangular inequality constraints, we first decompose the rotational inertia matrix as:
\begin{equation}
        \rotationalinertia = \rotationmatrix \principalmomentofinertiamatrix \rotationmatrix\T
\end{equation}
where $\rotationmatrix \in SO(3)$ is a rotation matrix, and $\principalmomentofinertiamatrix$ is a diagonal matrix containing the principal moments of inertia $\principalmomentofinertiascalar_1, \principalmomentofinertiascalar_2, \principalmomentofinertiascalar_3$.
The principal moments of inertia are all positive ($\principalmomentofinertiascalar_1 > 0$, $\principalmomentofinertiascalar_2 > 0$, $\principalmomentofinertiascalar_3 > 0$) such that $\rotationalinertia$ is positive definite. In addition to the positiveness of the principal moments of inertia, a physically realizable rotational inertia matrix $\rotationalinertia$ needs to have $\principalmomentofinertiamatrix$ that satisfies the triangular inequalities \citep{Wensing_LMI_2018, Traversaro_IdPhysicalConsistent_2016}:
\begin{equation}
        \principalmomentofinertiascalar_1 + \principalmomentofinertiascalar_2 \geq \principalmomentofinertiascalar_3 \mbox{, \quad}  \principalmomentofinertiascalar_2 + \principalmomentofinertiascalar_3 \geq \principalmomentofinertiascalar_1 \mbox{, \quad}  \principalmomentofinertiascalar_1 + \principalmomentofinertiascalar_3 \geq \principalmomentofinertiascalar_2
        \label{eq:triangle_ineq_principal_moment_of_inertia}
\end{equation}
In \cite{Wensing_LMI_2018, Traversaro_IdPhysicalConsistent_2016} the triangular inequality and $\rotationmatrix \in SO(3)$ constraints were encoded explicitly, here we propose a change of variables such that these constraints are encoded implicitly allowing us to utilize the standard gradient based optimizers of toolboxes such as PyTorch \citep{Paszke_PyTorch_AutoDiff_2017}.

We start out by introducing a set of unconstrained parameters $
\rotmataxisanglelearnableparams = \begin{bmatrix}
        \rotmataxisanglelearnableparamscalar_{1} &
        \rotmataxisanglelearnableparamscalar_{2} &
        \rotmataxisanglelearnableparamscalar_{3}
\end{bmatrix}\T$ that represent an axis-angle orientation from which the rotation matrix $\rotationmatrix$ can be recovered by applying the exponential map to the skew-symmetric matrix recovered from $\rotmataxisanglelearnableparams$.
    \begin{equation}
        \rotationmatrix = exp\Bigg(
        \begin{bmatrix}
            0    & -\rotmataxisanglelearnableparamscalar_{3} & \rotmataxisanglelearnableparamscalar_{2}\\
            \rotmataxisanglelearnableparamscalar_{3}  & 0    & -\rotmataxisanglelearnableparamscalar_{1}\\
            -\rotmataxisanglelearnableparamscalar_{2} & \rotmataxisanglelearnableparamscalar_{1}  & 0
        \end{bmatrix}
        \Bigg)
    \end{equation}
    where $exp(.)$ is the exponential mapping that maps $\rotmataxisanglelearnableparams$, a member of $so(3)$ group, to $\rotationmatrix$, a member of the $SO(3)$ group \citep{Murray_AMI_1994}.\\
    %
    Second, to satisfy triangular inequality constraints in Eq. \ref{eq:triangle_ineq_principal_moment_of_inertia} above, we can parameterize $\principalmomentofinertiascalar_1$, $\principalmomentofinertiascalar_2$, and $\principalmomentofinertiascalar_3$ as the length of the sides of a triangle. The length of the first 2 sides of the triangle are encoded by $\principalmomentofinertiascalar_1$ and $\principalmomentofinertiascalar_2$, and the length of the 3rd side is computed as 
    \begin{equation}
        \principalmomentofinertiascalar_3 = \sqrt{\principalmomentofinertiascalar_1^2 + \principalmomentofinertiascalar_2^2 - 2 \principalmomentofinertiascalar_1 \principalmomentofinertiascalar_2 \cos{\principalmomentofinertiaangle}}
    \end{equation}
    with $0 < \principalmomentofinertiaangle < \pi$.
    To encode that $\principalmomentofinertiascalar_1, \principalmomentofinertiascalar_2 > 0$, and $0 < \principalmomentofinertiaangle < \pi$ we choose the following parametrization:
    \begin{equation}
        \principalmomentofinertiascalar_1 = (\theta_{\sqrt{J_1}})^2 + \positivedefinitebias \mbox{, \quad} 
        \principalmomentofinertiascalar_2 = (\theta_{\sqrt{J_2}})^2 + \positivedefinitebias \mbox{, \quad} 
        \principalmomentofinertiaangle = \pi \textnormal{sigmoid}(\theta_a)
        \label{eq:principalmomentofinertiascalar_parameterization}
    \end{equation}
Thus, the learnable parameters of this parametrization are:
 \begin{equation}
        \triangprmlearnableparams = 
        \begin{bmatrix}
            \theta_{\sqrt{m}} &
            \masscenterofmasscombinedlearnableparam &
            \rotmataxisanglelearnableparamscalar_{1} &
           \rotmataxisanglelearnableparamscalar_{2} &
           \rotmataxisanglelearnableparamscalar_{3} &
           \theta_{\sqrt{J_1}} &
           \theta_{\sqrt{J_2}} &
           \theta_{a}
        \end{bmatrix}
        \label{eq:rotinertialearnableparams4}
    \end{equation}
    Note, even though the underlying learnable parameter vector $\triangprmlearnableparams$ in Eq. \ref{eq:rotinertialearnableparams4} is unconstrained during the parameter optimization via gradient descent, the intermediate parameters $\principalmomentofinertiascalar_1, \principalmomentofinertiascalar_2, \principalmomentofinertiascalar_3, \rotationmatrix$ always satisfy the hard constraints for physical consistency, i.e. that $\principalmomentofinertiascalar_1$, $\principalmomentofinertiascalar_2$, and $\principalmomentofinertiascalar_3$ are all positive and satisfy the triangle inequality constraints in Eq. \ref{eq:triangle_ineq_principal_moment_of_inertia}, as well as $\rotationmatrix \in SO(3)$. In other words, it \textit{always} lies within the constraint manifold during optimization.
    \AW{This paragraph rephrases some ideas already explained above and could probably be a lot more concise.}
\subsection{Covariance Parameterized Rotational Inertia Matrix (DiffNEA Cov)}
Alternatively, the triangular inequality constraint in Eq. \ref{eq:triangle_ineq_principal_moment_of_inertia} can be rewritten as \citep{Wensing_LMI_2018}:
%
\begin{equation}
        \densityweightedcovariancerotinertiamatrix = \frac{1}{2} Tr(\rotationalinertia) \eye_{3 \times 3} - \rotationalinertia \succ \zerovector
        \label{eq:def_linkdensityweightedcovariancerotinertiamatrix}
\end{equation}
which provides a somewhat easier and more intuitive representation. Again, in \cite{Wensing_LMI_2018} this constraint was imposed explicitly, through linear matrix inequalities. Here, we encode the constraint 
$\densityweightedcovariancerotinertiamatrix \succ \zerovector$ 
implicitly by enforcing a Cholesky decomposition 
plus a small positive bias $b$ on the diagonal: 
$\densityweightedcovariancerotinertiamatrix = \lowertriangularmatrix \lowertriangularmatrix\T + b \eye_{3 \times 3}$. 
We parametrize this lower triangular matrix $\lowertriangularmatrix$:
\begin{equation}
        \lowertriangularmatrix = 
        \begin{bmatrix}
            \theta_{\text{L}\Sigma_1} & 0 & 0\\
            \theta_{\text{L}\Sigma_4} & \theta_{\text{L}\Sigma_2} & 0 \\
            \theta_{\text{L}\Sigma_5} & \theta_{\text{L}\Sigma_6} & \theta_{\text{L}\Sigma_3}\\
        \end{bmatrix}
\end{equation}
and recover the rotational inertia matrix as:
\begin{equation}
        \rotationalinertia = Tr(\densityweightedcovariancerotinertiamatrix) \eye_{3 \times 3} - \densityweightedcovariancerotinertiamatrix
\end{equation}
with $Tr()$ is the matrix trace operation, and $\eye_{3 \times 3}$ is a $3 \times 3$ identity matrix. The learnable parameters of this parametrization are:
 \begin{equation}
        \covprmlearnableparams = 
        \begin{bmatrix}
            \theta_{\sqrt{m}} &
            \masscenterofmasscombinedlearnableparam &
            \theta_{\text{L}\Sigma_1} &
            \theta_{\text{L}\Sigma_2} &
            \theta_{\text{L}\Sigma_3} &
            \theta_{\text{L}\Sigma_4} &
            \theta_{\text{L}\Sigma_5} &
            \theta_{\text{L}\Sigma_6}
        \end{bmatrix}
        \label{eq:rotinertialearnableparams5}
    \end{equation}
This parametrization also generates fully consistent inertial parameter estimates, like the previous parametrization, however it is less complex to implement.
\section{Experiments} 
\label{sec:learn_inv_dyn_experiments}

In this Section, we evaluate our Torch implementation of the Newton-Euler algorithm, with the parametrizations introduced in Section \ref{sec:learn_inv_dyn_structured_learning_rep_diff_rnea}. We study how the parametrizations affect convergence speed when training the parameters, and how well the dynamics generalize to unseen scenarios. Furthermore, we compare a spectrum of structured dynamics learning approaches, starting from an unstructured MLP, and semi-structured models like DeLaN, to our highly structured approach. We start out with simulation experiments and then provide real system results on Kuka iiwa7 robot.

%
\subsection{Simulation}
%
In simulation, we collect training data on a simulated KUKA IIWA environment in PyBullet \\
\citep{Coumans_PyBullet}, by tracking sine waves in each joint with the ground-truth inverse dynamics model. The sine waves have time periods of $[23.0, 19.0, 17.0, 13.0, 11.0, 7.0, 5.0]$ seconds in each joint, and amplitudes $[0.7, 0.5, 0.5, 0.5, 0.65, 0.65, 0.7]$ times the maximum absolute movement in each joint. All dynamics models are trained with this sine wave motion dataset. All feed-forward neural networks involved in MLP and DeLaN models have $[32, 64, 32]$ nodes in the hidden layer with $tanh()$ activation functions.


We perform each experiment with 5 different random seeds --- which affects the random initialization as well as the random end-effector goal position to be tracked during generalization tests --- and then compute the performance statistics with mean and standard deviation across these.
\subsubsection{Training Speed, Generalization Performance, and Effectiveness of Inverse Dynamics Learning}
\begin{table}[ht]
\centering
\begin{footnotesize}
    \begin{tabular}{ |c||c|c|c|c|c| }
        \hline
        \multirow{3}{*}{Model} & \multicolumn{3}{|c|}{Sine Tracking (NSME)} & \multicolumn{2}{|c|}{End-Effector Tracking (NSME)}\\
        \cline{2-6}
        & \# Training Epochs & $\jointposition$ Tracking & $\jointvelocity$ Tracking & $\cartesianposition$ Tracking & $\cartesianvelocity$  Tracking\\
        \hline
        \hline
        Ground Truth   & N/A       & 0.000 & 0.000           & 0.005$\pm$0.006 & 0.008$\pm$0.010 \\
        MLP            & 23$\pm$3  & 0.000 & 0.001           & 0.256$\pm$0.405 & 5.542$\pm$6.980 \\
        DeLaN          & 58$\pm$19 & 0.000 & 0.001           & 0.016$\pm$0.008 & 0.254$\pm$0.278 \\
        DiffNEA NoStr  & 61$\pm$53 & 0.000 & 0.005$\pm$0.006 & 0.005$\pm$0.006 & 0.037$\pm$0.057 \\
        DiffNEA Symm   & 8$\pm$4   & 0.000 & 0.000           & 0.005$\pm$0.006 & 0.011$\pm$0.011 \\
        DiffNEA SPD    & 2$\pm$1   & 0.000 & 0.000           & 0.005$\pm$0.006 & 0.008$\pm$0.010 \\
        DiffNEA Tri    & 2$\pm$1   & 0.000 & 0.000           & 0.005$\pm$0.006 & 0.008$\pm$0.010 \\
        DiffNEA Cov    & 2$\pm$1   & 0.000 & 0.000           & 0.005$\pm$0.006 & 0.008$\pm$0.010 \\
        \hline
    \end{tabular}
    \caption{\small Comparison between models trained to optimize $\inversemodelloss$ on the sine motion dataset from simulation, in terms of training speed, joint position ($\jointposition$) and velocity ($\jointvelocity$) tracking, and generalization performance: end-effector position ($\cartesianposition$) and velocity ($\cartesianvelocity$) tracking unseen end-effector reaching tasks.}
    \label{tab:sim_training_and_generalization_comparison}
\end{footnotesize}
\end{table}
We train each model until it achieves at most a normalized mean squared error (NMSE) of $0.1$ for all joints. We record total epochs of training required to reach that level of accuracy and store the model once this accuracy has been achieved. 
Next, we evaluate model on tracking a) the sine motion itself (which the parameters were fitted for), and b) on a series of $5$ operational space control tasks. 
For the second task, we use a velocity-based operational space controller as described in \citep{Nakanishi_OpSpaceControl}, and use the learned inverse dynamics model within that controller. The results for convergence speed and tracking performance are averaged across the $5$ random seeds and summarized in Table \ref{tab:sim_training_and_generalization_comparison}. 
We measure the tracking performance through NMSE, which is the mean squared tracking error normalized by the variance of the target trajectory. The better the tracking, the less the controller relies on the feedback component, and more on the inverse dynamics model prediction. The behavior becomes more compliant as the contribution of linear feedback goes down.

%
%
Table \ref{tab:sim_training_and_generalization_comparison}, shows that the parametrizations of $\spdprmlearnableparams, \triangprmlearnableparams, \covprmlearnableparams$ outperform the less constrained learning, by training faster as well as low NMSE. Moreover, we see that DiffNEA performs close to ground-truth in performance, and generalizes better than the unstructured MLP as well as the DeLaN model: MLP model oscillates significantly in the end-effector velocity ($\cartesianvelocity$) tracking, while DeLaN also oscillates mildly in $\cartesianvelocity$ tracking.

%
\subsubsection{Online Learning Speed}
\begin{wrapfigure}{r}{0.5\textwidth}
\vspace{-1.5cm}
    \centering
      \caption{\small Online learning, with NMSE in log scale.}
    \includegraphics[width=0.95\linewidth]{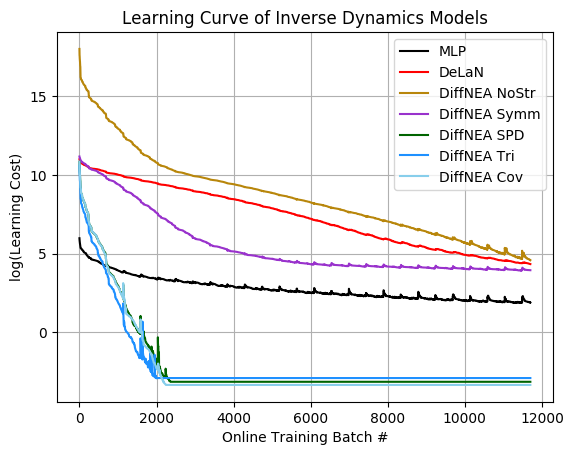}
    \label{fig:sim_online_learning_curve}
\vspace{-0.5cm}
\end{wrapfigure}

Next, we measure how fast each model can learn in an online learning setup. We train each model sequentially \textbf{without shuffling} on the sine motion data, where each batch is of size of 256. As the model trains with the sequential data, we measure its prediction performance through the NMSE on the entire dataset.

In Fig. \ref{fig:sim_online_learning_curve} we see that the DiffNEA model with rotational inertia $\rotationalinertia$ parameterized with symmetric positive definite (SPD) matrix, triangular parameterization, and covariance parameterization learns the fastest, but also generalizes the best to the yet unseen training data, outperforming other models in the online learning setup in simulation.
\FM{I don't understand how we have 12000 epochs of batches of 256 data point? we collect only 600*240 data points?}

\subsection{Real Robot Experiments}
For the real KUKA IIWA robot, we collect sine wave tracking data for about 240 seconds at 250 Hz using the default URDF (Unified Robot Description Format) parameters, and the Pinocchio C++ library \citep{pinocchioweb} for dynamics and kinematics of the robot. We noticed un-modeled friction dynamics not present in simulation, and added a joint viscous friction/damping model to both DeLaN and DiffNEA models, whose parameters are also learned from data. We use one positive constant per joint damping with parameterization similar to $\principalmomentofinertiascalar_1, \principalmomentofinertiascalar_2$ in Eq. \ref{eq:principalmomentofinertiascalar_parameterization}, but with $\positivedefinitebias = 0$ because each joint damping constant can be $0$.

%
\subsubsection{Evaluation}

\begin{table}[ht]
\centering
\begin{footnotesize}
    \begin{tabular}{ |c||c|c|c|c|c| }
        \hline
        \multirow{3}{*}{Model} & \multicolumn{3}{|c|}{Sine Tracking (NMSE)} & \multicolumn{2}{|c|}{End-Effector Tracking (NMSE)}\\
        \cline{2-6}
        & \# Training Epochs & $\jointposition$ Tracking & $\jointvelocity$ Tracking & $\cartesianposition$ Tracking & $\cartesianvelocity$  Tracking\\
        \hline
        \hline
        Default Model   & N/A & 0.001 & 0.009 & 0.000 & 0.016 \\
        MLP            & 2 & 0.000 & 0.011 & 0.003 & 0.513 \\
        DeLaN          & 4 & 0.001 & 0.013 & Unstable & Unstable \\
        DiffNEA Symm  & 3 & 0.001 & 0.012 & 0.000 & 0.013 \\
        DiffNEA SPD   & 3 & 0.001 & 0.013 & 0.000 & 0.014 \\
        DiffNEA Tri   & 2 & 0.001 & 0.013 & 0.000 & 0.012 \\
        DiffNEA Cov   & 2 & 0.001 & 0.012 & 0.000 & 0.015 \\
        \hline
    \end{tabular}
    \caption{\small Comparison between models trained to optimize $\inversemodelloss$ on the sine motion dataset on the real robot, in terms of the number of training epochs required to reach convergence, sine motion joint position ($\jointposition$) tracking, sine motion joint velocity ($\jointvelocity$) tracking, and generalization performance: end-effector position ($\cartesianposition$) and velocity ($\cartesianvelocity$) tracking NMSE on an end-effector tracking task (unseen task/situation during training).}
    \label{tab:rbt_training_and_generalization_comparison}
    \end{footnotesize}
\end{table}
%
%
Both MLP and DeLaN models converge to average training NMSE less than 0.1, while the DiffNEA models converge to average training NMSE 0.35, down from the default model with average NMSE 0.74.
However, as can be seen in Table \ref{tab:rbt_training_and_generalization_comparison}, the trained DeLaN model is unstable during the end-effector tracking task, while the trained MLP model has a large end-effector velocity ($\cartesianvelocity$) tracking NMSE due to oscillations. On the other hand, trained DiffNEA models still perform reasonably, showing its better generalization capability.
We attribute the imperfect training of DiffNEA models to unmodelled dynamics of the real system, such as static friction.

\GSutanto{Giovanni is still working on improving this real robot experiments until the deadline date.}

\section{Conclusion and Future Work} 
\label{sec:learn_inv_dyn_conclusion}
In this paper, we incorporate physical constraints in learned dynamics by adding structure to the learned parameters. This enables us to learn the dynamics of a robot manipulator in a computational graph with automatic differentiation, while keeping the learned dynamics physically plausible. We evaluate our approach on both simulated and real 7 degrees-of-freedom KUKA IIWA arm.
Our results show that the resulting dynamics model trains faster, and generalize to new situations better than other state-of-the-art approaches. 

We also observe that moving to the real robot creates new sources of discrepancies between the rigid body dynamics and the true dynamics of the system. Factors such as static friction cannot be sufficiently captured by our model, and point to interesting future directions.



\bibliography{references}

\end{document}